# Fundamental Matrix Estimation: A Study of Error Criteria

Mohammed E. Fathy[a,*], Ashraf S. Hussein[a], Mohammed F. Tolba[a]

[a]*Faculty of Computer and Information Sciences, Ain Shams University, Abbassia, 11566, Cairo, Egypt*

**Abstract**

The fundamental matrix (FM) describes the geometric relations that exist between two images of the same scene. Different error criteria are used for estimating FMs from an input set of correspondences. In this paper, the accuracy and efficiency aspects of the different error criteria were studied. We mathematically and experimentally proved that the most popular error criterion, the symmetric epipolar distance, is biased. It was also shown that despite the similarity between the algebraic expressions of the symmetric epipolar distance and Sampson distance, they have different accuracy properties. In addition, a new error criterion, Kanatani distance, was proposed and was proved to be the most effective for use during the outlier removal phase from accuracy and efficiency perspectives. To thoroughly test the accuracy of the different error criteria, we proposed a randomized algorithm for Reprojection Error-based Correspondence Generation (RE-CG). As input, RE-CG takes an FM and a desired reprojection error value $d$. As output, RE-CG generates a random correspondence having that error value. Mathematical analysis of this algorithm revealed that the success probability for any given trial is $1 - (2/3)^2$ at best and is $1 - (6/7)^2$ at worst while experiments demonstrated that the algorithm often succeeds after only one trial.

## 1. Introduction

The fundamental matrix (FM) relating two images $(I, I')$ is estimated from a number of correspondences between $I$ and $I'$. A correspondence is a pair of points $(p, p')$ on the two images $(I, I')$ that are believed to be projections of the same 3D point. Automatic algorithms for identifying correspondences not only introduce errors in the computed locations of the points (localization errors), but also produce totally false matches (outliers) (Zhang and Kanade, 1998). To get acceptable results, FM estimation usually starts by removing these outliers. Then, a one-step FM estimation technique such as the eight-point algorithm (Hartley and Zisserman, 2004; Zhang and Kanade, 1998) is used to obtain a better estimate of the FM by taking into account the effect of all the inliers rather than just a small, 7-point subset. Finally, the result obtained by the one-step method is refined using an iterative technique. FM error criteria play a vital role in the process of the FM estimation. An FM error criterion is a real-valued function that measures the amount of deviation of a given correspondence from the epipolar constraint parameterized by a given FM. FM error criteria are used in three different situations:

1. During the outlier removal phase of the FM estimation, an error criterion is used as a *distance function* to measure the proximity of each correspondence to the current FM hypothesis.

*Corresponding author. Tel.: +20 14 537 0021; fax: +20 2 2682 7574.

*Email addresses:* m.e.fathy (at) gmail.com (Mohammed E. Fathy), ashrafh (at) acm.org (Ashraf S. Hussein), fahmytolba (at) gmail.com (Mohammed F. Tolba)

2. During the iterative refinement of the FM, an error criterion is used as a *cost function* to be minimized over the space of rank-2 matrices.
3. An error criterion can also be used as an *accuracy measure* to compare different solutions obtained by different estimation methods.

The evaluation of the FM error criteria can be thought of as a primitive operation that is performed many times throughout the process of the FM estimation. Assuming that the outlier removal phase performs $L$ iterations over a set of $N$ correspondences that consist of $n$ inliers and $N-n$ outliers, the error criterion is evaluated $LN$ times. Also, if the iterative refinement runs $M$ iterations, the error criterion is evaluated at least $Mn$ times. If the iterative refinement phase uses a 7-tuple parameterization of the FM and forward difference is used to estimate the Jacobian of the residual vector, the error criterion needs to be evaluated $7Mn$ additional times. Hence, the evaluation of the error criterion should be as efficient as possible. This is more crucial when it is desired to compute the FM for each pair of images in an unorganized collection of thousands of images as in (Snavely et al., 2008). In this paper, the different FM error criteria were studied from both the accuracy and efficiency perspectives. We mathematically and experimentally proved that the most popular error criterion, the symmetric epipolar distance, is biased and that its accuracy properties are different from those of Sampson distance despite the similarity of the algebraic expressions of both criteria. In addition, we proposed a new error criterion, Kanatani distance, and it was shown experimentally that it is the most effective criterion for use during the outlier removal phase. We evaluated the accuracy and efficiency of the different error criteria in approximating the gold standard error criterion, the reprojection error, using different values of the reprojection error. To permit this evaluation, we introduced a new randomized algorithm for RE-based Correspondence Generation (RE-CG). RE-CG takes a reprojection error value and an FM at input and produces at output a randomly generated correspondence having that reprojection error value. The RE-CG algorithm is mathematically analyzed and it is shown that the success probability for any given trial is $1-(2/3)^2$ at best and $1-(6/7)^2$ at worst. Experiments revealed that the algorithm usually succeeds after just one trial. The remainder of this paper is organized as follows. The relevant related work is reviewed in Section 2. The mathematical notations adopted in this paper are presented in Section 3. Section 4 presents the existing FM error criteria. Section 5 mathematically proves that the symmetric epipolar distance (SED) is biased and that SED is dissimilar from Sampson distance despite the similarity between their algebraic expressions. Section 6 proposes the Kanatani distance criterion. Section 7 proposes the RE-CG algorithm and analyzes its success probability. The experimental evaluations are described and their results are discussed in Section 8. Finally, the research is concluded in Section 9.

## 2. Related Work

The algebraic distance (R) was the first error criterion to appear in the FM literature (Hartley, 1992; Olsen, 1992; Shashua, 1992). It was used to estimate the FM from a number of noisy correspondences by minimizing the sum of squares of algebraic distances ($\sum_i R_i^2$). The advantage of this formulation is that the cost function to be minimized is quadratic in the FM coefficients and hence can be solved in one step (linear least-squares problem) (Forsyth and Ponce, 2002; Hartley and Zisserman, 2004; Zhang and Kanade, 1998). Luong et al. (1993) showed that R is biased. As alternatives, they introduced the symmetric epipolar distance (SED) and also adapted the Sampson distance (RE1) for use in the FM estimation. In their FM estimation experiments over synthetic data, they concluded that the cost functions $\sum_i \text{SED}_i^2$ and $\sum_i \text{RE1}_i^2$ give similar results but superior to those obtained by minimizing $\sum_i R_i^2$ using the eight point algorithm. However, the way they utilized in their experiments to evaluate the difference between the exact (ground truth) FM and the estimated FM was not precise enough and so their conclusions are arguable. This is because the way they used to evaluate the difference between two FMs measured only the difference between the exact and estimated locations of the epipoles while ignoring the differences introduced by the other 3 parameters of the FM. A seemingly better way of measuring the difference between FMs was used by Zhang and Kanade (1998) and it was found that the cost function defined by SED gives slightly inferior results to those obtained by RE1 and RE. Their experiment was performed on a *single* stereo pair of real images where the ground truth FM was known a priori. Based on this result alone, Hartley and Zisserman (2004) advised to avoid using SED as cost functions for FM iterative refinement. Torr and Murray (1997) found in their experiments that RE1 agrees with RE for up to the fourth or fifth significant digit. However, they did not report whether this is true however large RE is. In this paper, we verified that RE1 starts to lose accuracy as RE exceeds 100 pixels. The similarity between the algebraic expressions of RE1 and SED was observed in different studies (Luong et al., 1993; Torr and Murray, 1997; Torr et al., 1998). This observation led to the belief that RE1 and SED are similar. However, we mathematically proved that $SED^2 \geq 2RE1^2$ which means that SED and RE1 give dissimilar values. In addition, the experimental results obtained in this paper confirm that SED and RE1 behave differently. For example, our experimental results show that SED tends to over-estimate RE whereas RE1 tends to under-estimate RE for large values of RE. For small values of RE, SED over-estimates RE whereas RE1 well-estimates RE. Being immediately physically intuitive, SED has been the most popular error criterion in use in prominent computer vision libraries such as OpenCV (Bradski and Kaehler, 2008; OpenCV: Open Computer Vision Library, 2009), VxL (VxL, 2009), and Bundler (Snavely et al., 2008). Many recent books in computer vision still present the iterative refinement of the FM using SED (Faugeras et al., 2001; Forsyth and Ponce, 2002; Ma et al., 2004). Furthermore, most of the comparative studies used SED as an accuracy measure to compare the accuracy of the FMs estimated by different methods (Armangué and Salvi, 2003; Forsyth and Ponce, 2002; Hartley and Zisserman, 2004). However, we mathematically and experimentally proved that $SED^2$ is a biased estimator of the gold standard error measure $RE^2$. Consequently, the values obtained by $SED^2$ should not be used to judge the relative accuracy of different FM estimates.

## 3. Mathematical Notations

In the rest of this paper, the homogeneous coordinate vector of a 2D Euclidean point $p = (\ x\ \ y\ )^T$ is denoted by $\tilde{p} = (\ x\ \ y\ \ 1\ )^T$. The point $\tilde{p}$ lies on the homogeneous line $l = (\ l_1\ \ l_2\ \ l_3\ )^T$ if it satisfies the line equation $l^T\tilde{p} = \tilde{p}^T l = l_1 x + l_2 y + l_3 = 0$. The function $n(l) = (\ l_1\ \ l_2\ )^T$ gives the inhomogeneous coordinates of the vector normal to $l$, whereas $u(l) = R(-\frac{\pi}{2})n(l) = (\ l_2\ \ -l_1\ )^T$ gives the inhomogeneous coordinates of the vector tangent to $l$. The signed distance $d(q, l)$ of a point $q = (\ x\ \ y\ )^T$ from a line $l$ is given by

$$d(q,l) \ = \frac{l^T\tilde{q}}{|n(l)|} \ = \frac{l^T\tilde{q}}{\sqrt{l_1^2 + l_2^2}} \quad (1)$$

The reader is referred to (Fathy, 2010; Hartley and Zisserman, 2004) for more details about homogeneous representations. Let $I$ and $I'$ be two perspective images of the same scene, $F$ be the 3x3 fundamental matrix (FM) relating $I$ and $I'$, and $\tilde{p}$ and $\tilde{p}'$ be the image projections of some 3D point $P$ on $I$ and $I'$. The epipolar constraint can then be written as follows

$$\tilde{p}^T F \tilde{p}' = 0 \quad (2)$$

The epipolar constraint function R: $\mathbf{R}^4 \to \mathbf{R}$ can also be written in the following equivalent ways:

$$\begin{aligned} R(B) \ &= R(p, p') = \ R(x, y, x', y') & (3)\\ &= \tilde{p}^T F \tilde{p}' & (4)\\ &= \tilde{p}^T l = \ l_1 x + l_2 y + l_3 & (5)\\ &= l'^T \tilde{p}' = \ l'_1 x' + l'_2 y' + l'_3 & (6) \end{aligned}$$

where $l = F\tilde{p}'$ is the epipolar line in $I$ that corresponds to $\tilde{p}'$ and $l' = (\tilde{p}^T F)^T = F^T\tilde{p}$ is the epipolar line in $I'$ that corresponds to $\tilde{p}$ (Forsyth and Ponce, 2002; Hartley and Zisserman, 2004). The epipolar constraint may be interpreted in a number of different ways. These interpretations provide the bases for some FM error criteria such as SED and RE. It may be interpreted as a line constraint that tests the incidence of the point $p'$ on the line $l' = F^T\tilde{p}$ or the incidence of the point $p$ on the line $l = F\tilde{p}'$. This is the point-on-line interpretation of the epipolar constraint. It may also be interpreted as testing the incidence of the correspondence $B = (x, y, x', y') = (p, p') \in \mathbf{R}^4$ on the hyper-surface $S$ implicitly defined by the epipolar constraint $\mathrm{R}(B) = 0$ (we call $S$ as the epipolar hyper-surface). This is the point-on-surface interpretation of the epipolar constraint. The reader is referred to (Faugeras et al., 2001; Forsyth and Ponce, 2002; Hartley and Zisserman, 2004) for more details about epipolar geometry.

## 4. Existing FM Error Criteria

The most-widely used error criteria are reviewed in this section. We started by describing the gold standard criterion, which is the reprojection error (RE). Next, the algebraic distance (R), the symmetric epipolar distance (SED), and Sampson distance (RE1) are presented. In the following discussion, it is assumed that $(p_i, p'_i)$ is the correspondence whose error is to be evaluated, $F$ is the fundamental matrix, and $l = F\tilde{p}'_i$ and $l' = F^T \tilde{p}_i$ are the homogeneous coordinate vectors of the epipolar lines in the two images.

### 4.1. Reprojection Error (RE)

The reprojection error (RE) is defined as the orthogonal distance in the 4D space defined by $(x, y, x', y')$ between the correspondence $(p_i, p'_i)$ and the 4D quadric S implicitly defined by the equation $\tilde{u}^T F \tilde{v}' = 0$. It is computed by solving the following optimization problem:

$$\mathrm{RE}_i^2 = \min_{\hat{p}_i, \hat{p}'_i} d^2(p_i, \hat{p}_i) + d^2(p'_i, \hat{p}'_i) \quad (7)$$
$$\text{subject to} \quad \hat{\tilde{p}}_i^T F \hat{\tilde{p}}'_i = 0$$

RE measures the minimum distance needed to bring $(p_i, p'_i)$ into perfect correspondence. The perfect correspondence $(\hat{p}_i, \hat{p}'_i)$ that minimizes (7) is called the optimally-corrected correspondence. It can be computed using Hartely-Sturm triangulation (Hartley and Sturm, 1997). Assuming the errors in the four measured coordinates $(x, y, x', y')$ are independent and identically-distributed random variables from a zero-mean Gaussian distribution, the optimally-corrected points $(\hat{p}_i, \hat{p}'_i)$ are the maximum likelihood estimate of the true correspondence $(\bar{p}_i, \bar{p}'_i)$ (Hartley and Zisserman, 2004; Kanatani et al., 2008). As a result, RE is regarded as the perfect error criterion and is considered the gold standard (Hartley and Zisserman, 2004). Finding the rank-2 FM that minimizes the sum of squares of reprojection errors is possible using non-linear least squares methods such as Levenberg-Marquardt (Hartley and Zisserman, 2004; Zhang and Kanade, 1998). However, Hartley-Sturm triangulation involves finding the roots of a polynomial of degree 6, making RE relatively expensive to compute (Hartley and Sturm, 1997; Hartley and Zisserman, 2004; Kanatani et al., 2008). If we take into consideration the frequency in which the error criterion is evaluated, it is necessary to find efficient, yet accurate, alternatives to RE. In this paper, we used RE as a gold standard against which other error criteria were evaluated.

### 4.2. Algebraic Distance (R)

The algebraic distance or epipolar error is the simplest criterion. It is defined as

$$\mathrm{R}_i = \mathrm{R}(p_i, p'_i) = \tilde{p}_i^T F \tilde{p}'_i \quad (8)$$

The algebraic distance does not have a physical meaning. It is proportional to the distance of a point from the epipolar line defined by the corresponding point:

$$\tilde{p}_i^T F \tilde{p}'_i = \lambda d(p_i, l) = \lambda' d(p'_i, l') \quad (9)$$

where $l = F\tilde{p}'_i$ and $l' = F^T \tilde{p}_i$ are the homogeneous coordinate vectors of the epipolar lines in the two images, and $\lambda = \sqrt{l_1^2 + l_2^2}$ and $\lambda' = \sqrt{l_1'^2 + l_2'^2}$. It was proved that $\lambda$ and $\lambda'$ introduce a bias and tend to bring the epipoles towards the image center (Luong et al., 1993; Zhang and Kanade, 1998). Due to this reason, it is used neither as a distance function for the outlier removal phase nor as an accuracy measure. The eight-point FM estimation method employs the sum of squared algebraic distances $\sum_i \mathrm{R}_i^2$ as a cost function where the solution is forced to be rank-2 in a post-processing step (Forsyth and Ponce, 2002; Hartley and Zisserman, 2004; Zhang and Kanade, 1998). Even though it is not very accurate, the eight-point method has the advantage of being a one-step method. So, it is usually used after outlier-removal to obtain an initial estimate of the FM that takes into account the whole set of inliers rather than just a seven-point sample. This initial estimate is then fed into an iterative refinement procedure to obtain a more accurate solution by minimizing a more accurate error criterion.

*4.3. Symmetric Epipolar Distance (SED)*

An alternative error measure that attempts to eliminate the bias introduced by the algebraic distance is the symmetric epipolar distance SED:

$$\begin{aligned} \text{SED}_i^2 &= d^2(p_i, l) + d^2(p_i', l') \\ &= \left( \frac{1}{l_1^2 + l_2^2} + \frac{1}{l_1'^2 + l_2'^2} \right) R_i^2 \end{aligned} \qquad (10)$$

It measures the geometric distance of each point to its epipolar line. Being immediately physically intuitive, this is the most widely used error criterion in practice during the outlier removal phase (OpenCV: Open Computer Vision Library, 2009; Snavely et al., 2008; VxL, 2009), during iterative refinement (Faugeras et al., 2001; Forsyth and Ponce, 2002; Snavely et al., 2008), and in comparative studies to compare the accuracy of different solutions (Armangué and Salvi, 2003; Forsyth and Ponce, 2002; Hartley and Zisserman, 2004; Torr and Murray, 1997). Besides being physically intuitive, SED has the merit of being efficient to compute. However, we proved in Section 5.1 that SED provides a biased estimate of the gold standard criterion RE.

*4.4. Sampson Distance (RE1)*

The Sampson distance RE1 is given by

$$\begin{aligned} \text{RE1}_i^2 &= \text{R}_i^2 / |\nabla \text{R}_i|^2 \\ &= \frac{1}{l_1^2 + l_2^2 + l_1'^2 + l_2'^2} \text{R}_i^2 \end{aligned} \qquad (11)$$

RE1 provides a first-order approximation of RE (Luong et al., 1993; Torr and Zissermann, 1997; Torr and Murray, 1997). It measures the distance between the correspondence $(p_i, p_i')$ and its Sampson correction (Torr and Zissermann, 1997). Although the algebraic expressions of RE1 and SED look similar (Luong et al., 1993; Torr and Murray, 1997), we theoretically and experimentally proved that they have quite different accuracy properties in Sections 5.2 and 8.2, respectively. Although less popular than SED, RE1 has been used as a distance function during the outlier removal phase (Kovesi, 2009; Torr and Murray, 1997) and as an accuracy measure for comparing different methods (Torr and Murray, 1997). In addition to the Levenberg-Marquardt formulation, minimization of $\sum_i \text{RE1}_i^2$ as a cost function has been formulated as an iterative eigenvalue problem using the Fundamental Numerical Scheme (FNS) and its constrained variant (CFNS) (Chojnacki et al., 2004; Zhang and Kanade, 1998). The (parameterized) Levenberg-Marquardt and the CFNS formulations yield a rank-2 matrix, whereas the FNS does not necessarily yield a rank-2 matrix. So, the result of FNS must be post-processed to be rank-2. Torr and Murray experimentally showed that the values obtained by RE1 agree with the values obtained by RE on up to the 4th or even the 5th significant digit (Torr and Murray, 1997). However, they did not mention whether this is true for relatively small as well as large values of RE. This is important since if RE1 deviates from RE for large values, using RE1 during the outlier removal phase will be less recommended. In Section 8, the accuracy of RE1 for small values as well as large values of RE is experimentally evaluated.

## 5. SED Relations to Other Criteria

In this section, we theoretically established the relation between SED and the gold standard criterion RE and its first-order approximation RE1. In particular, we showed that $\text{SED}^2$ is a biased estimator of $\text{RE}^2$ in Section 5.1. In addition, we confirmed that despite the similarity of the algebraic expressions of $\text{SED}^2$ (10) and $\text{RE1}^2$ (11), there is an inequality between $\text{SED}^2$ and $\text{RE1}^2$ similar to the inequality established between $\text{SED}^2$ and $\text{RE}^2$ in Section 5.1. In the following discussion, it is assumed that $F$ is the fundamental matrix, $B_i = (p_i, p_i') \in \mathbf{R}^4$ is the noisy correspondence whose error is to be evaluated, $A_i = (\hat{p}_i, \hat{p}'_i)$ is the closest perfect correspondence (i.e. optimal correction) to $B_i$, and $l = F\tilde{p}'_i$ and $l' = F^T \tilde{p}_i$ are the homogeneous coordinate vectors of the epipolar lines corresponding to $p'_i$ and $p_i$, respectively.

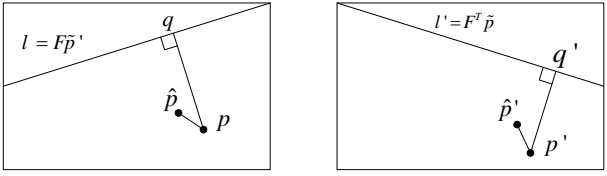

Figure 1: Symmetric epipolar distance SED is given by $\text{SED}^2 = d^2(p, q) + d^2(p', q')$. Reprojection error RE is given by $\text{RE}^2 = d^2(p, \hat{p}) + d^2(p', \hat{p}')$.

### 5.1. SED and RE

Despite the popularity of $\text{SED}^2$ in computer vision software libraries and literature, we mathematically proved that $\text{SED}^2$ provides a biased estimate of $\text{RE}^2$. Since $B_i = (p_i, p'_i)$ is noisy, $p_i$ does not lie on the epipolar line $l$. One way to correct $B_i$ is to fix $p'_i$ and replace $p_i$ by any other point lying on $l$. Suppose we correct $B_i = (p_i, p'_i)$ to $C_i = (q_i, p'_i)$, where $q_i$ is the orthogonal projection of $p_i$ onto $l$, as shown in Fig. 1. It follows that $\tilde{q}_i^T l = \tilde{q}_i^T F \tilde{p}'_i = 0$ and so the correspondence $C_i$ lies on the epipolar hyper-surface $S$. Since $A_i$ is the closest correspondence in $S$ to $B_i$, it follows that

$$d^2(B_i, C_i) \geq d^2(B_i, A_i) \qquad (12)$$

$$d^2(p_i, q_i) + d^2(p'_i, p'_i) \geq \text{RE}_i^2 \qquad (13)$$

$$d^2(p_i, l) \geq \text{RE}_i^2 \qquad (14)$$

An alternative way to correct $B_i = (p_i, p'_i)$ is to fix $p_i$ and replace $p'_i$ by any other point on $l'$. Suppose we correct $B_i = (p_i, p'_i)$ to $D_i = (p_i, q'_i)$ where $q'_i$ is the orthogonal projection of $p'_i$ onto $l'$. So, $l'^T \tilde{q}'_i = \tilde{p}_i^T F \tilde{q}'_i = 0$ and it follows that the correspondence $D_i$ lies on the epipolar hyper-surface $S$. Thus,

$$d^2(B_i, D_i) \geq d^2(B_i, A_i) \qquad (15)$$

$$d^2(p_i, p_i) + d^2(p'_i, q'_i) \geq \text{RE}_i^2 \qquad (16)$$

$$d^2(p'_i, l') \geq \text{RE}_i^2 \qquad (17)$$

From (14) and (17), we have

$$d^2(p_i, l) + d^2(p'_i, l') \geq 2\text{RE}_i^2 \qquad (18)$$

$$\text{SED}_i^2 \geq 2\text{RE}_i^2 \qquad (19)$$

$$\text{SED}_i^2 = 2\text{RE}_i^2 + 2b_i^2 \rightarrow \frac{\text{SED}_i^2}{2} = \text{RE}_i^2 + b_i^2 \qquad (20)$$

where $b_i^2$ is a bias term. Our experimental results showed that $b_i^2$ is a non-constant whose value depends on the coordinates of the correspondence $B_i$ and the FM $F$. That is, if $M_i$ and $M_j$ are two correspondences having the same squared reprojection error $\text{RE}^2$, $\text{SED}^2$ may prefer $M_i$ to $M_j$ by giving $M_i$ a higher bias $b_i^2$ than the bias $b_j^2$ which is given to $M_j$. Therefore, estimating the FM by minimizing the cost function $\sum_i \text{SED}_i^2$ may lead to sub-optimal (i.e. biased) results. During the outlier removal phase, correspondences that have the same $\text{RE}^2$ with respect to the same FM hypothesis should be given the same classification as inliers or outliers. If SED is adopted as the distance function during the outlier removal phase, SED may penalize some of the correspondences by giving them higher bias than the remaining correspondences even though they have the same reprojection error (RE). In turn, this will lead to inconsistent classification of some of the correspondences as outliers (i.e., false outliers). Finally, SED should not be used as an accuracy measure to compare the accuracy of different solutions, since SED may penalize one solution by giving it an overall bias that is greater than the overall bias given to the other solutions.

### 5.2. SED and RE1

Even though the algebraic expressions of $\text{SED}^2$ and $\text{RE1}^2$ look similar (Luong et al., 1993; Torr and Murray, 1997) at first glance, it should not be concluded that they have the same accuracy properties. A more thorough investigation of their algebraic expressions reveals that there is a discrepancy between $\text{SED}^2$ and $\text{RE1}^2$ similar to the discrepancy that exists between $\text{SED}^2$ and $\text{RE}^2$ as follows. Since

$$l_1^2 + l_2^2 + l_1'^2 + l_2'^2 \geq l_1^2 + l_2^2 \qquad (21)$$

it follows that

$$\frac{1}{l_1^2 + l_2^2} \geq \frac{1}{l_1^2 + l_2^2 + l_1'^2 + l_2'^2} \quad (22)$$

Similarly, we have

$$\frac{1}{l_1'^2 + l_2'^2} \geq \frac{1}{l_1^2 + l_2^2 + l_1'^2 + l_2'^2} \quad (23)$$

From (22) and (23), we have

$$\frac{1}{l_1^2 + l_2^2} + \frac{1}{l_1'^2 + l_2'^2} \geq \frac{2}{l_1^2 + l_2^2 + l_1'^2 + l_2'^2} \quad (24)$$

Multiplying both sides by $0.5R^2$ yields

$$\frac{1}{2}\left(\frac{1}{l_1^2 + l_2^2} + \frac{1}{l_1'^2 + l_2'^2}\right)R^2 \geq \frac{1}{l_1^2 + l_2^2 + l_1'^2 + l_2'^2}R^2 \quad (25)$$

Referring to (10) and (11), we can finally rewrite (25) as follows

$$\frac{\text{SED}^2}{2} \geq \text{RE1}^2 \quad (26)$$

Besides the discrepancy between $\text{SED}^2$ and $\text{RE1}^2$ indicated by (26), our experimental results indicated another aspect of the difference between $\text{SED}^2$ and $\text{RE1}^2$: while $\text{SED}^2$ tends to over-estimate $\text{RE}^2$ for all values of $\text{RE}^2$, $\text{RE1}^2$ well-estimates $\text{RE}^2$ for correspondences having small values of $\text{RE}^2$, and under-estimates $\text{RE}^2$ with correspondences having large values of $\text{RE}^2$. This emphasizes that $\text{SED}^2$ and $\text{RE1}^2$ have quite different accuracy properties.

## 6. Kanatani Distance (REK)

Recently, Kanatani et al. have proposed an iterative triangulation technique (Kanatani et al., 2008). It uses an iterative scheme to approximate the optimally-corrected points $(\hat{p}, \hat{p}')$. The corrected correspondence can then be fed into a one-step triangulation technique such as the Direct Linear Transform (DLT) to find the coordinate vector of the space point that generates this correspondence (Hartley and Zisserman, 2004; Kanatani et al., 2008). Kanatani's iterative correction scheme can be used to define a new error criterion which we call Kanatani distance (REK). REK proceeds by approximating the optimally-corrected points $(\hat{p}, \hat{p}')$ using the iterative correction scheme, and then measuring the squared distance between $(\hat{p}, \hat{p}')$ and $(p, p')$ in the same way as equation (7). Since one iteration of Kanatani's correction scheme is equivalent to Sampson correction (Kanatani et al., 2008), it follows that REK is equivalent to RE1 if we apply only one iteration of the iterative scheme. Further iterations of the iterative scheme should give more accurate results than those obtained by RE1. Our experimental results assert that REK is the most accurate criterion to use during the outlier removal phase. Even though it is iterative, REK exhibits an adaptive nature. It increases the number of iterations for large values of the reprojection error in order to maintain the accuracy. Our experiments demonstrated that, for small values of the reprojection error, REK executes just a few iterations to converge. As the reprojection error increases, the number of iterations taken by REK also increases, but at a lower rate. Correspondences with coordinates corrupted with zero-mean Gaussian noise and with standard deviations up to 10 pixels were used to test the accuracy of Kanatani's iterative scheme (Kanatani et al., 2008). The results obtained by Kanatani's iterative correction scheme agreed, in terms of accuracy, with those obtained by Hartley-Sturm triangulation. However, Kanatani's iterative correction scheme took significantly lower processing time than that taken by Hartley-Sturm triangulation (Kanatani et al., 2008). For the purpose of triangulation, it is sufficient to test with standard deviations up to 10 pixels since it is assumed that outliers have been removed before the triangulation phase. If we wish to use REK during the outlier removal phase, we must study the accuracy aspects of REK when the noise levels exceed 10 pixels. This is because the FM defined by a particular random subset of the correspondences might not fit well all the available correspondences. Consequently, some of the error levels involved are expected to be rather high. We took this into our consideration so that the experiments tried very large ($10^6$ pixels) as well as very small ($10^{-6}$ pixels) values of the reprojection error.

# 7. RE-Based Correspondence Generation (RE-CG)

Later on, we shall test the relative accuracy of the different error criteria at different levels of RE. So, we must be able to specify a desired value of RE and randomly generate a noisy correspondence $B = (p, p')$ that has that value of RE. The algorithm described below takes at input a desired RE value (say $d$), the parameters of two cameras $(C, C')$, and the corresponding FM $F$. The algorithm returns a random noisy correspondence $B$ such that $\text{RE}(B) = d$, where RE is measured with respect to the FM $F$ given at the input. For $m$ trials, the algorithm has a success probability of at least $1 - (6/7)^{2m}$. In practice, the algorithm succeeds after just one trial.

## 7.1. The Randomized Algorithm

The correspondences that satisfy the epipolar constraint (perfect correspondences) can be thought of as the points in the 4D space, defined by the coordinates $(x, y, x', y')$, that lie on the hyper-surface $S$ implicitly defined by

$$R(p, p') = \tilde{p}^T F \tilde{p}' = 0 \qquad (27)$$

This way, we can restate our problem as that of generating a 4D point whose distance from the quadric $S$ is $d$. Our solution to this problem starts by randomly generating a perfect correspondence $A = (\hat{p}, \hat{p}')$ for which $R(A) = 0$. This can be performed by randomly generating a 3D point $P$ in front of both cameras and finding its images on both cameras. Alternatively, we can use the parametric form of the epipolar hyper-surface proposed in Section 7.2. Then, we find if there is a correspondence $B = (p, p')$ whose distance from the hyper-surface $S$ is $d$ and that has $A$ as the nearest point on $S$. If $B$ exists, it follows that $\overrightarrow{AB}$ must be orthogonal to $S$ at $A$ and that $\left|\overrightarrow{AB}\right| = d$:

$$B - A = \pm d \hat{\nabla} R(A) \qquad (28)$$

where

$$\begin{aligned}
\hat{\nabla} R(\hat{p}, \hat{p}') &= \frac{1}{|\nabla R(\hat{p}, \hat{p}')|} \nabla R(\hat{p}, \hat{p}') \\
&= \frac{1}{\sqrt{\left|n(\hat{l})\right|^2 + \left|n(\hat{l}')\right|^2}} \begin{pmatrix} n(\hat{l}) \\ n(\hat{l}') \end{pmatrix} \\
&= \frac{1}{\sqrt{\hat{l}_1^2 + \hat{l}_2^2 + \hat{l}_1'^2 + \hat{l}_2'^2}} \begin{pmatrix} \hat{l}_1 \\ \hat{l}_2 \\ \hat{l}_1' \\ \hat{l}_2' \end{pmatrix} \quad (29)
\end{aligned}$$

$\hat{l}_1$ and $\hat{l}_2$ are the first two coordinates of the epipolar line $\hat{l} = F\tilde{\hat{p}}'$ passing through $\hat{p}$. Similarly, $\hat{l}_1'$ and $\hat{l}_2'$ are the first two coordinates of the epipolar line $\hat{l}' = F^T \tilde{\hat{p}}$ passing through $\hat{p}'$. So, the two points $B_1 = A + d\hat{\nabla}R(A)$ and $B_2 = A - d\hat{\nabla}R(A)$ are computed and RE is evaluated at each point using Hartley-Sturm correction as described in section 4.1. If one of them has $\text{RE} = d$, the algorithm returns it. If both of them have smaller REs, the trial fails and the algorithm makes a new trial starting from a different perfect correspondence generated randomly. We found, in practice, that this strategy converges in most of the cases after the first trial. We provided an explanation of this phenomenon here. Given the noisy correspondence $B$, the nearest perfect correspondence is obtained by solving the following optimization problem:

$$\begin{aligned}
\underset{C}{\text{argmin}} \quad & d^2(B, C) \\
\text{subject to} \quad & R(C) = 0 \qquad (30)
\end{aligned}$$

By Lagrange's theorem for constrained optimization, it is easy to see that the correspondence $A$ is a critical correspondence of the optimization problem (30). Given the correspondence $B = (p, p')$ as input, the Hartley-Sturm correction scheme solves a more constrained version of the optimization problem (30). Rather than considering all the correspondences lying on $S$, it just considers every correspondence $C = (q, q')$ from $S$ with the property that $(q, q')$ are the orthogonal projections of $(p, p')$ on the epipolar lines $l_q = F\tilde{q}'$ and $l_{q'} = F^T\tilde{q}$, respectively. If we prove that $A = (\hat{p}, \hat{p}')$ satisfies the additional constraints introduced by the Hartley-Sturm optimization, it will follow that $A$ is one of the critical points of the Hartley-Sturm optimization as it is a critical point of the more general optimization problem (30). Substituting (29) into (28), we obtain

$$\begin{pmatrix} p - \hat{p} \\ p' - \hat{p}' \end{pmatrix} = \pm \frac{d}{|\nabla R(\hat{p}, \hat{p}')|} \begin{pmatrix} n(\hat{l}) \\ n(\hat{l}') \end{pmatrix} \quad (31)$$

where $\hat{l} = F\tilde{p}'$ and $\hat{l}' = F^T\tilde{p}$ are the epipolar lines passing through $\hat{p}$ and $\hat{p}'$, respectively. It follows that $\overrightarrow{\hat{p}p}$ is orthogonal to $\hat{l}$. So, $\hat{p}$ is the orthogonal projection of $p$ onto $\hat{l}$. Similarly, $\hat{p}'$ is the orthogonal projection of $p'$ onto $\hat{l}'$. It follows that the correspondence $A$ lies in the domain of Hartley-Sturm's correction optimization. Since $A$ is a critical point of the more general optimization problem given in (30), it follows that $A$ is one of the critical points considered by Hartley-Sturm's optimization. Since the Hartley-Sturm algorithm identifies these critical points by finding the roots of a polynomial of degree six (Hartley and Sturm, 1997; Hartley and Zisserman, 2004; Kanatani et al., 2008), there are either 2, 4, or 6 real critical points, with $A$ being one of these points. Beside these critical points, Hartley-Sturm optimization considers an additional point determined by an asymptotic analysis of the function being optimized(Hartley and Sturm, 1997; Hartley and Zisserman, 2004; Kanatani et al., 2008). So, there are 3 possible cases $\{C_3, C_5, C_7\}$ where the number of candidate solutions in case $C_n$ is $n$. As the probability of occurrence of each case in $\{C_3, C_5, C_7\}$ is unknown, we just consider the success probability in each case separately. For a case involving $n$ candidates, we assume that all such candidates are equally likely to be the optimal solution. In this case, the failure probability is $(n-1)/n$. As described earlier, each trial of the algorithm involves two sub-trials (one with $B_1$ and another with $B_2$). The best situation occurs when each sub-trial involves the fewest possible number of candidates. The failure probability of the trial in such a case is $(2/3)(2/3) = (2/3)^2$ and thus the success probability is $1 - (2/3)^2$. In the worst case, each sub-trial involves 7 candidates. The success probability in this case is $1 - (6/7)^2$. If the algorithm performs $m$ trials, failure happens when all the $2m$ sub-trials fail. So, the success probability is $1 - (2/3)^{2m}$ at best and $1 - (6/7)^{2m}$ at worst.

### 7.2. A Parametric Form for The Epipolar Hyper-surface

For very large values of RE, we have found in practice that the number of trials, needed by the algorithm to succeed, increases. To solve this problem, we found that choosing a perfect correspondence, such that its distance from the epipoles is proportional to the specified RE value $d$, retains the success rate of the algorithm. It is not clear how to control how far the generated correspondence is from the epipoles by simply generating a 3D point and projecting it. So, we have proposed an alternative parameterization of the epipolar hyper-surface that makes it possible to control the distance of the generated correspondence from the epipoles. Assuming that the two epipoles do not lie at infinity, the parameterization is given by

$$\begin{aligned} C(t, d, d') &= \begin{pmatrix} p(t, d) \\ p'(t, d') \end{pmatrix} \\ &= \begin{pmatrix} e + d\hat{u}(l(t)) \\ e' + d'\hat{u}(l'(t)) \end{pmatrix} \end{aligned} \quad (32)$$

The parameter $t$ selects the pair of corresponding epipolar lines $(l(t), l'(t))$ onto which the generated correspondence lies. The parameters $(d, d')$ control the distances of the two generated points from the corresponding epipoles $(e, e')$. The vector function $\hat{u}(l)$ gives the unit vector tangent to the line $l$. It is defined as:

$$\hat{u}(l) = \frac{1}{|u(l)|} u(l) \quad (33)$$

The epipolar line $l(t)$ is defined as the line passing through the epipole $a_1 = \tilde{e}$ and the point $a_2(t) = \tilde{e} + \begin{pmatrix} cos(t) & sin(t) & 0 \end{pmatrix}^T$. The coordinate vector $l(t)$ is found by forming the cross product $l(t) = a_1 \times a_2(t)$. The coordinate vector $l'(t)$ is simply computed as $l'(t) = F^T \tilde{p}(t, 1)$. In case the first epipole $e$ is at infinity, we define $p(t, d)$ simply as $ti + de$ if $e$ and $i$ are linearly independent, where $i = \begin{pmatrix} 1 & 0 \end{pmatrix}^T$. Otherwise, we define it as $tj + de$, where $j = \begin{pmatrix} 0 & 1 \end{pmatrix}^T$. In case the second epipole $e'$ is at infinity, we evaluate $p'(t, d)$ as the 2D point equivalent to the point with homogeneous coordinate vector $\left( \begin{pmatrix} e'^T & 0 \end{pmatrix}^T \times l'(t) \right) + d' \begin{pmatrix} e'^T & 0 \end{pmatrix}^T$.

Equation (32) makes it possible to explicitly control the distance of the generated pair of points from the epipoles by providing suitable values for the parameters $d$ and $d'$. To generate a random correspondence with this equation, we generate $t$ from the uniform distribution $U(-\pi, \pi)$, and $d$ and $d'$ from the zero-mean Gaussian distribution with standard deviation 1000 RE. We found, in practice, that this leads to a high success rate.

## 8. Experiments

Two experiments are conducted. The first experiment tests the success rate of the RE-CG randomized algorithm while the second experiment compares the performance of the three error criteria SED, RE1, and REK against the gold standard, RE. The experiments were conducted on a Dell Vostro notebook equipped with an Intel Core 2 Duo T7500 (2.2 GHz) processor and 2.0 GB RAM. All computations were performed in double precision. The experiments generate pairs of random cameras. Each camera is parameterized by a position $P$, an orientation $R$ ($R$ is the rotation matrix that transforms the world axes so that they are parallel to the axes of the camera), a focal length $f$, and a principal point $(u, v)$. The frame of the first camera is always taken coincident with the world frame. That is, the experiments set $P = 0$ and $R = I$. The position $P'$ of the second camera is selected randomly on the unit sphere centered at the origin. Assuming that $U(a, b)$ is the uniform distribution of the real numbers in $[a, b]$, the generation of $P'$ is done by generating two random angles $s$ and $t$ from the distributions $U(-90, 90)$ and $U(0, 360)$, respectively, and setting $P' = (\cos s \cos t, \sin s, \cos s \sin t)$. The orientation $R'$ of the second camera is obtained by generating 3 random angles $\theta$, $\phi$, and $\psi$ from the distributions $U(-135, 135)$, $U(-90, 90)$, and $U(0, 360)$, respectively, and setting $R'$ as the product of the elementary rotations $R_y(\theta) R_x(\phi) R_z(\psi)$. Assuming that $N(a, b)$ is the normal distribution with mean $a$ and standard deviation $b$, the focal lengths of the cameras are taken from $N(f_{avg} = 1300, \sigma = 250)$. The principal point $(u, v)$ of each camera is obtained by taking $u$ and $v$ from $N(u_{avg} = 399.5, \sigma = 133.33)$ and $N(v_{avg} = 299.5, \sigma = 100)$, respectively. The experiments generate random 3D points in a cube $C = [-3 \times 10^5, 3 \times 10^5]^3$. That is, each coordinate belongs to the interval $[-3 \times 10^5, 3 \times 10^5]$.

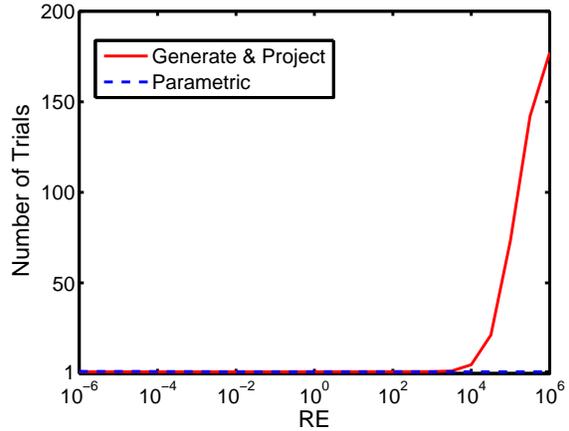

Figure 2: For each value of RE (on the x-axis), the plot shows the mean number of trials taken by RE-CG to succeed.

### 8.1. RE-CG Success Rate

This experiment practically tests the success rate of the proposed randomized algorithm RE-CG. Given a value of RE, the experiment generates two random cameras $(C, C')$, and runs RE-CG twice: the first with perfect correspondences being generated by generating and projecting 3D points, while the other with perfect correspondences being generated by using the parametric form of the epipolar hyper-surface proposed in section 7.2. The number of trials performed to reach success by each of these two variants of RE-CG is recorded. If a given algorithmic variant makes 200 trials (arbitrary number) without success, it terminates with failure. The experiment is performed for different values of RE. At each value of RE, the experiment is repeated 1000 times and the average as well as the standard deviation of the num-bers of trials made by each algorithmic variant are computed. Figure 2 graphs the mean number of trials each algorithmic variant makes against RE. Both algorithmic variants have the mean number of iterations fixed at one as long as RE < 1000. Beyond this, only the parametric variant maintains an average of one trial. Figure 3 graphs the standard deviation of the number of trials each algorithmic variant makes against RE. For all values of RE, the number of trials made by the parametric variant has zero deviation from the mean (which is one trial). The deviation of the other variant becomes non-zero when RE exceeds 1000. Let the value of RE at which the performance of the Generate & Project (GP) variant starts to degrade be denoted by $d_v$. For the current experimental settings, $d_v = 1000$ pixels. The particular value of $d_v$ depends on two factors. The first is the values of the focal lengths used in the generation of the random cameras. Indeed, increasing $f_{avg}$ amplifies the value of $d_v$. When the experiment was repeated with $f_{avg} = 13,000$ (which is 10 times larger), the value of $d_v$ increased to about $10^4$ (which is also 10 times larger). The other factor is how far the projection of the cube $C$ lies from the epipoles. Although $d_v$ can be increased by choosing $C$ in such a way that it projects far enough from the epipoles, it is not clear how this can be done. The natural solution is to generate the correspondence directly in the image space where it is easy to control the distance of the generated correspondence from the epipoles. Thus, RE-CG often succeeds after the first trial using both algorithmic variants when the desired value of RE is relatively small. Beyond this, only the parametric variant of the algorithm retains the ability to achieve success after one iteration.

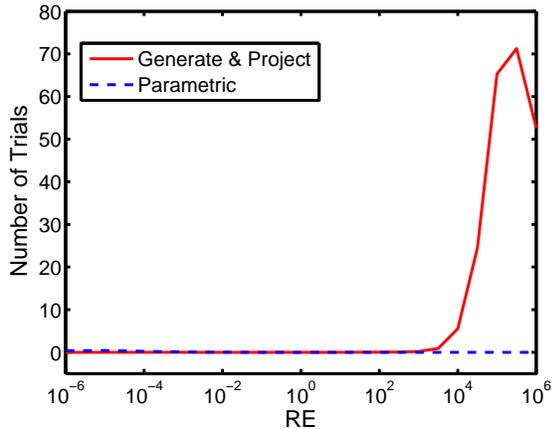

Figure 3: For each value of RE (on the x-axis), the plot shows the standard deviation of the numbers of trials taken by RE-CG to succeed.

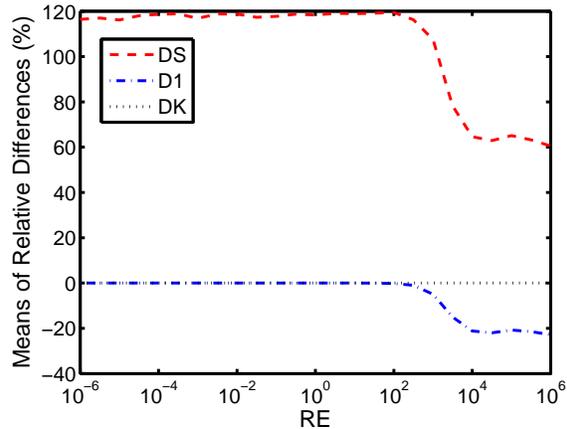

Figure 4: For each value of RE (on the x-axis), the plot shows the means relative differences between the error criteria and RE.

8.2. FM Error Criteria

The experiment studies the nature of the bias term introduced by SED. The experiment also measures the accuracy and efficiency of RE1 and REK in approximating the gold standard RE for different levels of reprojection error (RE). For every level $d$ of RE, the experiment generates two random cameras $(C, C')$, and a random correspondence $(p, p')$ that has a reprojection error $d$ using the RE-CG algorithm. The experiment then evaluates each of the tested error criteria (RE, SED, RE1, and REK) at the generated correspondence $(p, p')$ and records the following

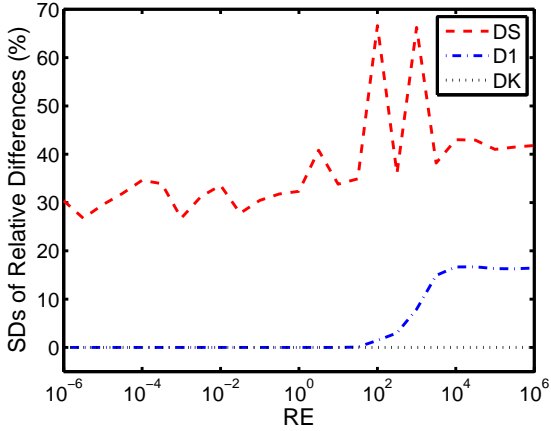

Figure 5: For each value of RE (on the x-axis), the plot shows the standard deviations of the relative differences between the error criteria and RE.

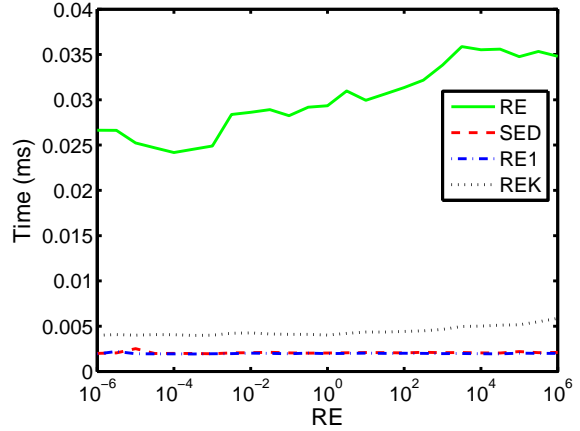

Figure 6: For each value of RE (on the x-axis), the plot shows the mean running times taken by the different error criteria.

measurements (DS, D1, DK, TE, TS, T1, TK, IK):

$$\text{DS} = \frac{0.5\text{SED}^2 - \text{RE}^2}{\text{RE}^2} * 100\% = \frac{b^2}{\text{RE}^2} * 100\% \qquad (34)$$

$$\text{D1} = \frac{\text{RE1}^2 - \text{RE}^2}{\text{RE}^2} * 100\% \qquad (35)$$

$$\text{DK} = \frac{\text{REK}^2 - \text{RE}^2}{\text{RE}^2} * 100\% \qquad (36)$$

$$(37)$$

DS, D1, and DK measure the relative difference (in percentage) between the error criteria and the gold standard RE. TE, TS, T1, and TK are the times taken to evaluate RE, SED, RE1, and REK, respectively. IK is the number of iterations taken by Kanatani's iterative correction scheme. We have set the maximum iteration count of Kanatani's iterative correction scheme to 1000 and its convergence constant $\delta$ to $10^{-6}$. We have also adjusted the convergence check in the implementation of Kanatani's scheme so that it switches the check from absolute to relative for large reprojection errors. That is, if the reprojection error at iteration $i$ is small ($E_i \leq 1$), we test the convergence by making the absolute check $|E_i - E_{i-1}| \leq \delta$. If the reprojection error is large ($E_i > 1$), we test for convergence by making the relative check $|E_i - E_{i-1}| \leq \delta E_i$. We use the polynomial root finder provided by the GNU Scientific Library (Gough, 2009) to solve the sixth-degree polynomial computed by the Hartley-Sturm correction scheme. For every level of reprojection error included in our test, we repeat the above procedure (i.e., generate a random correspondence and take measurements) 1000 times. Figure 4 plots for each value of reprojection error RE (on the x-axis) the average of the measured relative differences DS, D1, and DK. Figure 5 is a plot of the standard deviations of the relative differences DS, D1, and DK against the reprojection error RE. Figure 6 is a plot of the average running times of the different error criteria against the reprojection error RE. From these plots, we can make the following observations:

- It is apparent that the mean of the relative bias DS is always positive. This agrees with the theoretical relation between $\text{SED}^2$ and $\text{RE}^2$ we obtained in Section 5.1.

- It can be seen that the standard deviation of the relative bias DS introduced by SED is relatively large ($> 30\%$) at every value of RE. A non-zero standard deviation means that some or all of the measurements deviate from their mean. This means that SED gives widely different error values to correspondences that have the same reprojection error RE. The everywhere-inaccurate behavior of $SED^2$ is explained by the fact that $SED^2$ was originally derived to fix the problem of another criterion (the multiplicative bias of the algebraic distance) rather than to provide a good approximation to $RE^2$. As described in Section 5.1, this makes SED less recommended for use as a distance function in the outlier removal phase, as a cost function in the iterative refinement phase, or as an accuracy measure to compare different estimates of the FM.

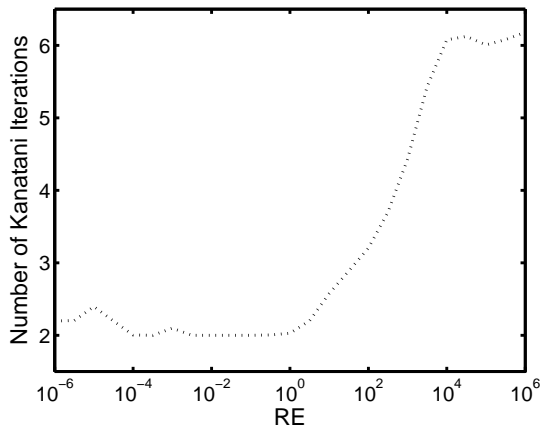

Figure 7: For each value of RE (on the x-axis), the plot shows the average number of iterations taken by REK.

- The mean and standard deviation of D1 remain close to zero for small to moderate values of RE. This means that RE1 provides an accurate estimate of RE as long as the correspondences have moderate values of RE. Since the running time of RE1 is always the smallest as apparent in Fig. 6, it is the best criterion to recommend from the accuracy and efficiency perspectives when it is not expected to have correspondences with large values of RE. This is the situation during the FM iterative refinement phase and the FM accuracy comparison phase because it is assumed that outliers have already been removed before these stages. On the other hand, the mean of D1 drops below zero for large values of RE ($> 100$ pixels). That is, RE1 tends to under-estimate RE when the correspondences that have large values of RE. In addition, the standard deviation of D1 increases when the reprojection error increases beyond 100 pixels. This means that RE1 gives less accurate estimates of RE for large values of RE. So, it is less accurate to use RE1 to approximate RE with correspondences having large values of RE, which is the situation during the outlier removal phase.

- Figures 4 and 5 reveal that $RE1^2$ starts to get inaccurate during the interval $I = [10^2, 10^4]$ of RE. The interval $I$ marks the transition from relatively small to relatively large values of RE. Both non-iterative criteria (RE1 and SED) undergo changes in behavior during that interval. Since RE is measured in pixels and since the focal length values express the number of pixels per physical distance unit on the image plane, there is some sort of dependence between the particular values of $I$ and the focal length values employed in the experiment. Indeed, when we repeated the experiment with 10 times larger focal length values, the interval $I$ increased to about $[10^3, 10^5]$.

- The mean and standard deviation of DK remain nearly zero for all values of RE. Figure 7 shows the number of iterations taken by Kanatani's iterative correction scheme for each value of RE. Although, the number of iterations and running time increase with the RE level, the running time TK remains below the running time TE taken by the gold standard RE (TK is always less than 0.2 TE). The ability of REK to automatically adapt the number of iterations based on the magnitude of the reprojection error and to provide an accurate estimate of RE, even for large values of RE, makes REK the most effective to use during the outlier removal phase. The results also show that Kanatani's iterative correction scheme is as accurate as Hartley-Sturm correction, even for very noisy correspondences. Meanwhile, Kanatani's iterative correction scheme is at least five times more efficient than Hartley-Sturm correction.

The deterioration in the accuracy of RE1 may be explained by the fact that RE1 uses a first-order approximation of the quadratic epipolar constraint. This approximation performs well as long as the distance $\text{dist}(B, A)$ between the noisy correspondence $B$ and its optimal correction $A$ is relatively small. As soon as $\text{dist}(B, A)$ (which equals RE by definition) increases, the first-order approximation becomes less accurate and the difference $d_1 = |\frac{D1}{100}| * \text{RE}^2$ between $\text{RE1}^2$ and $\text{RE}^2$ starts to increase. Although $d_1$ continues to increase as $\text{RE}^2$ increases, its growth rate $|\frac{D1}{100}|$ stops increasing when RE becomes very large. This is explained by the fact that the epipolar constraint (which is linearly approximated by RE1) is just a low-order (quadratic) polynomial.

## 9. Conclusions

The different FM error criteria have been presented and analyzed. It was mathematically proved that the popular SED is biased. In addition, it was experimentally verified that the bias value, introduced by SED, varies from a point to another, suggesting that the use of SED should be avoided. A new error criterion, Kanatani distance (REK), was proposed and it was experimentally found that it is the most effective error criterion for use as a distance function during the outlier removal phase of the FM estimation. The accuracy of Kanatani's iterative correction scheme was experimentally found equivalent to, yet five to seven times faster than, the accuracy of Hartley-Sturm correction, even for very noisy correspondences. Experiments demonstrated that Sampson distance (RE1) provides an accurate approximation for RE as long as the level of noise is limited. Since this is the situation after the outlier removal phase, the small runtime taken by RE1 promotes it to be the most suitable error measure for use as a cost function and as an accuracy measure. A new randomized algorithm was developed to generate random correspondences with pre-specified reprojection error values. The algorithm was mathematically analyzed and an estimate of the success probability for any given trial was shown to be $1 - (2/3)^2$ at best, and $1 - (6/7)^2$ at worst. Practical experiments showed that the algorithm succeeds often after just one trial.

## Acknowledgements

We would like to express our thanks to Peter Sturm for providing the source code of Hartley-Sturm triangulation and Yasuyuki Sugaya for making the implementation of Kanatani triangulation publicly available.